\documentclass[conference]{IEEEtran}
\IEEEoverridecommandlockouts
\PassOptionsToPackage{numbers}{natbib}
\usepackage{amssymb,amsthm}
\usepackage{amsmath}
\usepackage{graphicx}
\usepackage{adjustbox}
\usepackage{caption}
\usepackage{subcaption}
\usepackage{times}
\usepackage{fancyhdr,graphicx,amsmath,amssymb}
\usepackage[ruled,vlined,linesnumbered]{algorithm2e}
\usepackage[utf8]{inputenc} % allow utf-8 input
\usepackage[T1]{fontenc}    % use 8-bit T1 fonts
\usepackage{hyperref}       % hyperlinks
\usepackage{url}            % simple URL typesetting
\usepackage{booktabs}       % professional-quality tables
\usepackage{amsfonts}       % blackboard math symbols
\usepackage{nicefrac}       % compact symbols for 1/2, etc.
\usepackage{microtype}      % microtypography
\usepackage{svg}

\usepackage{cite}
\usepackage{amsmath,amssymb,amsfonts}
\usepackage{algorithmic}
\usepackage{graphicx}
\usepackage{textcomp}
\usepackage{xcolor}
\begin{document}

\title{Deep Spatial Domain Generalization}
% {\footnotesize \textsuperscript{*}Note: Sub-titles are not captured in Xplore and
% should not be used}
% \thanks{Identify applicable funding agency here. If none, delete this.}
% }

% \author{\IEEEauthorblockN{Anonymous Author}
% \IEEEauthorblockA{\textit{Anonymous Institute}}
% }

\author{\IEEEauthorblockN{Dazhou Yu, Guangji Bai, Yun Li, Liang Zhao}
\IEEEauthorblockA{\textit{Department of Computer Science} \\
\textit{Emory University}\\
Atlanta, USA \\
\{Dazhou.Yu, Guangji.Bai, Yun.Li, Liang.Zhao\}@emory.edu}

}

\maketitle

\begin{abstract}
Spatial autocorrelation and spatial heterogeneity widely exist in spatial data, which make the traditional machine learning model perform badly. Spatial domain generalization is a spatial extension of domain generalization, which can generalize to unseen spatial domains in continuous 2D space. Specifically, it learns a model under varying data distributions that generalizes to unseen domains.  Although tremendous success has been achieved in domain generalization, there exist very few works on spatial domain generalization. The advancement of this area is challenged by: 1) Difficulty in characterizing spatial heterogeneity, and 2) Difficulty in obtaining predictive models for unseen locations without training data. To address these challenges, this paper proposes a generic framework for spatial domain generalization. Specifically, We develop the spatial interpolation graph neural network \footnote{https://github.com/dyu62/Deep-domain-generalization} that handles spatial data as a graph and learns the spatial embedding on each node and their relationships. The spatial interpolation graph neural network infers the spatial embedding of an unseen location during the test phase. Then the spatial embedding of the target location is used to decode the parameters of the downstream-task model directly on the target location. Finally, extensive experiments on ten real-world datasets demonstrate the proposed method's strength.
\end{abstract}


\section{Introduction}
Traditional machine learning models are typically under the independent and identically distributed (i.i.d.) assumption, meaning the data samples are independent of each other and follow the same distribution. However, this assumption generally cannot be held for spatial data which have spatial autocorrelation and heterogeneity. Spatial autocorrelation makes the spatial location of a sample and corresponding spatial attributes informative and samples not independent and identically distributed (non-i.i.d.). Spatial heterogeneity includes spatial non-stationarity and spatial anisotropy. Spatial non-stationarity means that sample distribution varies across locations. Spatial anisotropy means that the spatial dependency between sample locations is non-uniform along different locations.
% Specifically, the air pollution index value (e.g., PM2.5) of a location is not only affected by the independent factors on this location but also affected by the nearby location's pollution index value. 
Specifically, the air pollution concentration of a location is usually a complex function of various independent variables but the relative importance of the independent variables are changing with locations, 
e.g., the population density and distances from emissions sources play an essential role in PM2.5 pollution concentration in Urban built-up areas. But in rural areas, the relative humidity is greatly attributed to the diffusion of PM2.5.
This requires us to have some customization on different models in different locations. However, in the training set, we usually only have observations from a limited number of locations. Hence, it is prevalent that we need to execute prediction tasks in locations unseen in the training set. This results in a very challenging task where we need to predict the model in a new location without any training data. This paper focuses on this new problem which we call spatial domain generalization, which is a spatial extension of domain generalization  \cite{muandet2013domain}. 

Domain generalization learns a model under varying data distributions that generalizes to unseen domains. It is derived from and goes beyond domain adaptation, which builds the bridge between source and target domains by characterizing the transformation between the data from these domains \cite{ben2010theory}. Current domain generalization only covers domains with categorical indices \cite{muandet2013domain} or time sequential domains \cite{nasery2021training} but has not covered spatial domains which require considering unique problems such as spatial autocorrelation and spatial heterogeneity. Another thread of research comes from the spatial data mining area, where people propose techniques such as Geographically weighted regression (GWR) \cite{wheeler2010geographically} to handle spatial heterogeneity. Most of the time, prescribed models are used where the underlying spatial distribution and correlation need to be presumed and predefined by the model designer which may not reflect the true spatial process that is usually complex and unknown. Especially, these models only consider distances and ignore other spatial information such as direction. What's more, these models share the feature extractor on all locations and only generate different coefficients in the last layer so they cannot capture complex heterogeneity within data.

The spatial domain generalization is challenged by several critical bottlenecks, including 
\textbf{1) Difficulty in characterizing spatial heterogeneity.} The data distribution is not identical in the entire space and is changing with respect to locations' confounding and characteristics. A simple global model cannot explain the relationships between variables. So the nature of the model must alter over space to reflect the structure within the data. Modeling the spatially changing relationships requires making the model location-sensitive. Feeding the coordinate values as part of input features is intuitive. However, such a method cannot leverage the fact of the other features' dependency on location and  other confounding factors varying among locations. It is necessary yet difficult to quantitatively figure out how the spatial heterogeneity impacts the models while there is no "one-fits-all" rule for it. It is highly imperative yet challenging to have some techniques that can automatically learn from the data. 
\textbf{2) Difficulty in obtaining predictive models for unseen locations without training data.}
Due to the spatial heterogeneity, the local models in different locations can be very different in order to capture the relationships between predictors and the target variable. When training data is not provided in some locations, the method must have the capacity to generalize to these unseen locations. This is as difficult as zero-shot learning. 
% While it is easy to get many samples in a fixed location, building a dense observer network is expensive. The sensors are sparsely distributed in the space and can provide labels for a small portion of domains. 

In order to  address the above challenges, we propose a generic framework for deep spatial domain generalization, which generates the predictive models for any unseen spatial domains. More specifically, to address the first challenge, we propose a novel spatial interpolation graph neural network (SIGNN) to learn the spatial embedding of each location and the relationships between them in the training set and infer the spatial embedding of unseen locations during the test phase. The spatial embedding of the target location is then used to decode the parameterized model directly without training data on the target location. This solves the second challenge. 
Our contribution includes
\begin{itemize}
  \item \textbf{We propose a framework for spatial domain generalization.} The framework doesn't assume the data distribution and learns the spatial embeddings for all the locations in the training set in an end-to-end manner. It is also compatible with general predictive task models such as regression models and multi-layer perceptrons (MLP).
  \item \textbf{We develop the spatial interpolation graph neural network.} It handles spatial data as a graph and uses the edge representation to learn the spatial embedding on each node and their relationships by doing graph convolution operations. It also interpolates the spatial embedding at any location so our method can generalize to unseen locations. 

\item \textbf{We conduct extensive experiments.} We validated the efficacy of our method on ten real-world datasets for classification and regression tasks. Our method outperforms state-of-the-art models on most of the tasks.
\end{itemize}

\section{Related Work}
In this section, we summarize the works in the field of domain adaptation and domain generalization.
Machine learning systems often assume that training and test data follow the same distribution, which, however, usually cannot be satisfied in practice. Domain Adaptation (DA) aims to build the bridge between source and target domains by characterizing the transformation between the data from these domains ~\cite{ben2010theory,ganin2016domain,tzeng2017adversarial}. Domain Adaptation (DA) has received great attention from researchers in the past decade~\cite{ben2010theory,ganin2016domain,tzeng2017adversarial}. Under the big umbrella of DA, continuous domain adaptation considers the problem of adapting to target domains where the domain index is a continuous variable (temporal DA is a special case when the domain index is 1D). Approaches to tackling such problems can be broadly classified into three categories: (1) biasing the training loss towards future data via transportation of past data\cite{hoffman2014continuous}, (2) using time-sensitive network parameters and explicitly controlling their evolution along time~\cite{mancini2019adagraph}, (3) learning representations that are time-invariant using adversarial methods~\cite{wang2020continuously}. The first category augments the training data, the second category reparameterizes the model, and the third category redesigns the training objective. However, data may not be available for the target domain, or it may not be possible to adapt the base model, thus requiring Domain Generalization.

A diversity of DG methods have been proposed in recent years. According to~\cite{wang2021generalizing}, existing DG methods can be categorized into the following three groups, namely: (1) \emph{Data manipulation:} This category of methods focuses on manipulating the inputs to assist in learning general representations. There are two kinds of popular techniques along this line: a). Data augmentation~\cite{tobin2017domain}, which is mainly based on augmentation, randomization, and transformation of input data; b). Data generation~\cite{qiao2020learning}, which generates diverse samples to help generalization. (2) \emph{Representation learning:} This category of methods is the most popular in domain generalization. There are two representative techniques: a). Domain-invariant representation learning~\cite{ganin2016domain}, which performs kernel, adversarial training, explicitly features alignment between domains, or invariant risk minimization to learn domain-invariant representations; b). Feature disentanglement~\cite{li2017domain}, which tries to disentangle the features into domain-shared or domain-specific parts for better generalization. (3) \emph{Learning strategy:} This category of methods focuses on exploiting the general learning strategy to promote the generalization capability.

\section{Methodology}

\begin{figure*}[htb]
  \centering
  \includegraphics[width=\textwidth]{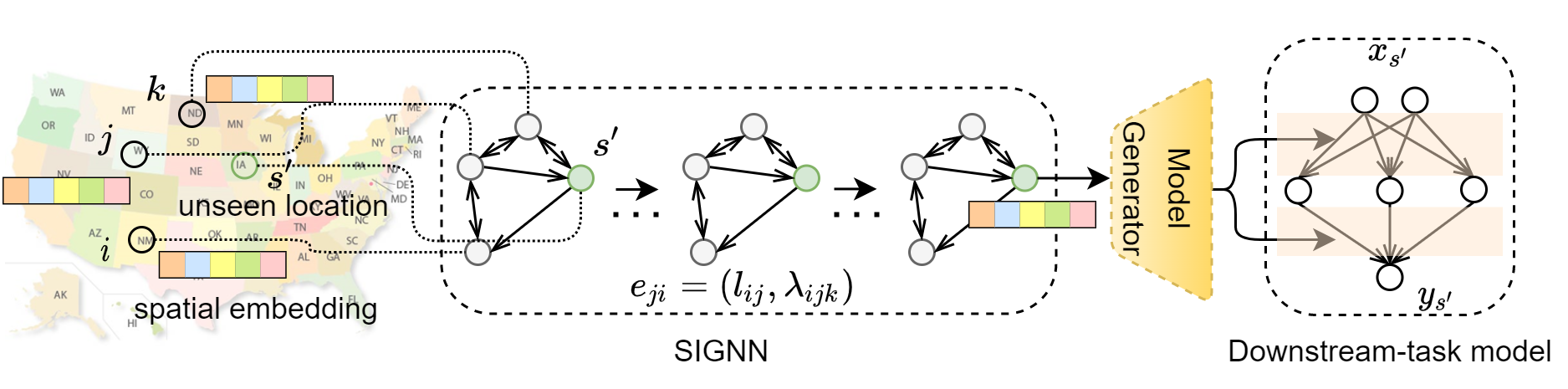}
%   \fbox{\rule[-.5cm]{0cm}{4cm} \rule[-0.5cm]{8cm}{0cm}}
  \caption{Illustration of the proposed framework. The unseen location's spatial embedding is interpolated by SIGNN. The edge representation contains both the distance and direction information. The spatial embedding is decoded to the weights of the downstream-task model.}
  \label{fig:arch}
\end{figure*}
In this section, we first provide the problem formulation and the challenges of the problem, then we introduce our proposed framework and how it solves the challenges.

\subsection{Problem formulation}\label{sec:pro_formu}
In this paper, we denote a geo-location by its 2D coordinate values $s\in \mathbb{R}^2$, and each $s$ is associated with a spatial domain $ (\mathcal{X}_s\times \mathcal{Y}_s)$, where we could have a set of samples $(\mathbf{x}_{s}, \mathbf{y}_{s}) =\{ (x_i,y_i)\in (\mathcal X_s\times \mathcal Y_s)\}_{i=1}^{N_s}$ where $x_i\in\mathcal X$ is $i$-th input sample from the domain $\mathcal X_s$, while $y_i\in\mathcal Y$ is the $i$-th output sample from the domain $\mathcal Y_s$. For the classification problem, $y_i$ can be further narrowed to a binary value. 

In opposition to an assumption that the relationship $f$ remains unchanged among dependent variables $x_i\in\mathcal X_s$ and independent variables $y_i\in\mathcal Y_s$  in the space $\mathbb{R}^2$, spatial heterogeneity describes a condition in which the relationships between some sets of variables $\{x_i,y_i\}$ are heterogeneous throughout space, i.e., $f_s \neq f_{s'}$ if $s \neq s'$. A static global model cannot capture the changes in relationships, thus Domain Generalization (DG) models which could reflect the heterogeneous relationships within the data play a vital role in spatial analysis. 

Our goal in this paper is to build a model that proactively captures the data concept drift across different geo-locations. Given a set of data samples $\{(\mathbf{x}_{s}, \mathbf{y}_{s})\}_{s\in S_0}$ from seen domains, where $S_0$ denotes the set of seen locations, we aim to learn the predictive mapping functions $f_s:\mathcal X_s\rightarrow \mathcal Y_s$ for downstream tasks such as classification or regression on location s. Here the location can either be seen  (i.e.,  $s\in S_0$) or unseen  (i.e., $s\in (\mathbb{R}^2-S_0)$). The former is spatial multitask learning while the latter is spatial domain generalization. Therefore, our problem is a generalization of both of them.

\subsection{Proposed Method}
\subsubsection{Spatial domain generalization}
We propose a bi-level framework as shown by Fig.  \ref{fig:arch} which generates the predictive models for any unseen spatial domains. Generally speaking, we propose a novel spatial interpolation graph neural network (SIGNN) to learn the target location's spatial embedding. The spatial embedding of the target location is then used to decode the parameterized model directly without training data on the target location. The general procedures of unseen domain generalization and model training are outlined in the following and detailed in Sections \ref{sec:compo1}.

\paragraph{Spatial $K$-nearest neighbor graph} For any location $s$ we will first build a spatial $K$-nearest neighbor graph upon $s$ and seen locations $S_0$ that is defined as $G(s,S_0;Z)= (V(s,S_0),E(s,S_0);Z)$, where node set $V(s,S_0)=S_0\bigcup \{s\}$ is just the union of the current location $s$ and seen locations $S_0$ defined before. So in the case that $s$ is a seen location, then $V$ is reduced to $S_0$.  $E(s, S_0) \subseteq V \times V$ denotes the relationships among all the locations, which will be detailed in Section \ref{sec:rep}. For simplicity, we omit the input and use $V$ and $E$ directly in the following. Let $\mathcal{N}_i^{(K)}$ denote node $v_i$'s $K$-nearest-neighbors, the nodes whose Euclidean distance from $v_i$ is less than or equal to the $k$-th largest Euclidean distance between any node and $v_i$.  To be specific, for a node $v_j \in \mathcal{N}_i^{(K)}$, a directed edge $ (v_j, v_i)$ exists from $v_j$ to $v_i$, so there are exactly $K$ nodes pointing to $v_i$. $Z= \{z_s\}_{s\in{S_0}}$ denotes the spatial embeddings for all the locations except the current location $s$, namely $S_0-\{s\}$, where $z_s$ is the spatial embedding vector for location $s$. Here the spatial embeddings are also the node features.

\paragraph{Unseen domain model generation} When doing spatial domain generalization, we are interested in generating the predictive model for an unseen location $s'\in\mathbb{R}^2-S_0$. And the spatial embedding for location $s'$ is spatially interpolated by our SIGNN via our newly proposed spatial interpolation graph convolutions $a (s';E,Z)$ by referring to the spatial embeddings of all seen spatial locations $S_0$. Then the spatial embedding of $s'$ is fed into the model generator to generate the parameterized function $f_{s'}$, namely the downstream task's model with the following function
\begin{equation}
    f_{s'}=d_{\varphi} (g_{\theta} ({s'};{E,Z})),
\end{equation}
where $d_{\varphi}$ denotes the downstream-task-model generator parameterized by ${\varphi}$, $g_{\theta}$ denotes SIGNN parameterized by ${\theta}$.
The downstream task can be any classification or regression task on location $s'$ such as weather classification, air pollution prediction, and so forth. We will elaborate on the details of transferring $a$'s output to a specific task's model in Section \ref{sec:compo1}.

\paragraph{Model Training} The above model generation for unseen location requires learning spatial embeddings $Z=\{z_s\}_{s\in{S_0}}$, model parameters $\theta$ of SIGNN $g_\theta$, and parameters $\varphi$ of model generator $d_\varphi$. In the following, we will introduce how to jointly learn them in the training phase. For each seen location, as mentioned in Section \ref{sec:pro_formu}, we know the input and output data of the downstream task. Hence our training objective is to maximize the likelihood given the prior of $p(Z)$,  by learning the unknown spatial embedding and model parameters, 

\begin{equation}
    \arg\max\limits_{Z,\varphi,\theta} \{ p (\mathbf{Y}|\mathbf{X},\varphi,\theta,Z)p (Z)\},
\end{equation}
the above equation is equal to minimizing the negative logarithm of the likelihood as follows,
\begin{equation}
    \arg\min\limits_{Z,\varphi,\theta} \{-\ln{p (\mathbf{Y}|\mathbf{X},\varphi,\theta,Z)}-\ln p (Z)\},
\end{equation}
where $\mathbf{Y} $ and $\mathbf{X}$ denote the prediction and input for all samples from all domains  ($\{\{ (x_i,y_i)\in (\mathcal X_s\times \mathcal Y_s)\}_i^{N_s}\}_{s\in S_0}$), respectively. Since $Z$ can be any continuous value, its prior distribution $p(Z)$ can be trivially assumed as an isotropic Gaussian normal distribution, we have
\begin{equation}\label{eq:p}
\arg\min\limits_{Z,\varphi,\theta} \{-\ln{p (\mathbf{Y}|\mathbf{X},\varphi,\theta,Z)}+\frac{1}{2}||Z||^2\}.
\end{equation}
Hence the first term is a  downstream task-specific prediction loss and the second term is a $\ell_2$  norm that regularizes $Z$. So the first term can also be more specifically expressed as $\sum_{s \in S_0} loss (f_{s} (\mathbf{x}_{s}),\mathbf{y}_{s})$,
where the parameter $W_s$ of each location $s$'s downstream predictive function $f_s$ is calculated as
\begin{equation}
    W_s=d_{\varphi}(z_s)=d_{\varphi}(g_{\theta}(s;E,Z)).
\end{equation}

In the following, we will more concretely introduce the prediction and model parameter training of our overall framework. Then in Section \ref{sec:compo1}, we will detail our SIGNN model and graph generator for generating the downstream-task model. Lastly, in Section \ref{sec:rep}, we will drill down into our edge representation.

\subsubsection{Unseen domain model generator}\label{sec:compo1}
In this subsection, we first introduce the details of our SIGNN model for doing the spatial embedding interpolation for unseen locations and then elaborate on the graph generator for generating the downstream task model using the interpolated spatial embedding.

\paragraph{Spatial interpolation graph neural network (SIGNN)}
As mentioned above, our SIGNN model $g_{\theta}(s;E,Z)$ aims at inferring the spatial embedding for a given location $s$, based on other locations $S_0$'s spatial embeddings and their spatial correlation with $s$. A key challenge unique to spatial interpolation beyond general message passing in graph neural networks is how to comprehensively represent such correlation among locations. Existing works that typically only consider the distances among the locations to represent their correlation  cannot consider the integrated spatial information such as the orientation of neighbors which are indispensable for spatial interpolation.

To achieve this, in our SIGNN we propose a novel edge representation $E(s,S_0)$ which is detailed in section \ref{sec:rep} and here we first introduce SIGNN and its convolutional operations based upon the edge representation and spatial embedding.

SIGNN is a stack of $M$ spatial interpolation graph convolutional layers $a_u, u=1,2,...,U$ , namely $g_{\theta}= a_U\circ a_{U-1}\circ\dots a_1$, where the input to each spatial interpolation graph convolutional layer is the target location, the set of our novel edge representations and spatial embeddings, namely $(s;E,Z)$. The spatial interpolation graph convolutional layer interpolates the spatial embedding $z_s$ as its output while the edge representations remain the same for each layer.

In order to do the interpolation, the spatial interpolation graph convolutional layer $a_u$ generates a pairwise weight $\omega_{ji}^{(u)}$ for each node $v_i$ and its neighbors $v_j\in \mathcal{N}_i^{(K)}$, then the spatial embedding of each node is updated by calculating a weighted sum of the spatial embeddings of neighboring nodes, namely ${z}_i^{(u+1)}=\sum_{j=1}^{K} \omega_{ji}^{(u)}*{z}_{j}^{(u)}$, where $\omega_{ji}^{(u)}$ equals
\begin{equation*}
\frac{\text{exp} (\sigma (\vec{\alpha}^T[m_1 ({e}_{ji})||m_2 ({z}_i^{(u)})||m_2 ({z}_j^{(u)})]))}{\sum_{k, v_k\in \mathcal{N}_i^{(K)}}\text{exp} (\sigma (\vec{\alpha}^T[m_1 ({e}_{ik})||m_2 ({z}_i^{(u)})||m_2 ({z}_k^{(u)})]))},
\end{equation*}
where $e_{ji}\in E$ denotes the edge representation for edge $(v_j, v_i)$, $z_i^{(u)}$ and $z_j^{(u)}$ denote the spatial embedding of node $v_i, v_j$ at layer $a_u$ respectively, $m_1$ and $m_2$ denote two MLP models that augment the spatial embedding and edge representation respectively, $||$ denotes the concatenation operation, $\sigma$ denotes the nonlinear activation function LeakyRuLU, $\vec{\alpha}$ denotes a vector parameter that transforms the concatenated vector to a scalar. We also use the softmax function to normalize the weights. Finally, we select the spatial embedding $z_s$ for location $s$ as the output. 
\paragraph{Downstream-task model generator}
% The model generator $d_{\varphi}$ decodes the spatial embedding $z_s$ to the weights of downstream-task-model, namely $W_s=d_{\varphi}(z_s)$, which can also be seen as a  graph .

% and the goal is to find an optimal transform matrix to map the input vector into a separable space. 
 Many shallow models like linear regression, logistic regression, and support vector machines manipulate the input vector with matrix operations such as multiplication between input and weight vectors. Such matrix operation can be considered as the fully-connected layer or other types of layers, with or without a nonlinear activation function. When the models go deep, then multiple layers are stacked into deep neural networks. Hence each of all these shallow or deep models for location $s$ can be denoted as its parameter which is network structured. Here we can formally define such a network as $\mathcal{G}=(\mathcal{V},\mathcal{E}; W_s)$, where $\mathcal{V}$ are the neurons, $\mathcal{E}$ are the links between neurons, and $W_s$ are the link weights for the model at location $s$. Here the model parameter $W_s$ is namely the output of our model generator $d_{\varphi}: W_s=d_{\varphi}(z_s) $. To be specific, a neural network can be represented as an edge-weighted graph $\mathcal{G}$, where each node ${v} \in \mathcal{V}$ corresponds to a neuron and each edge ${e} \in \mathcal{E}$ corresponds to the connection weight between two neurons. Following works in \cite{bui2021spatial}, we use a three-layer MLP to generate the downstream-task model's weights $W_s$. Then the model can load the weights and perform the task.

% So these methods can also be seen as a graph-structured model. Therefore, the generated graph is general enough and any downstream-task model can be formulated as $\mathcal{G}=(\mathcal{V},\mathcal{E})$.

% Here we consider a simple case where the topology of the downstream task  model is fixed, but the weight between the neurons is changing across domains. 

% To discriminate a domain $s$ from other domains, the model generator $d_{\varphi}$ takes in the spatial embedding $z_s$ and generates a local model that captures the characteristics of the spatial domain.
\subsubsection{Edge representation for spatial interpolation}\label{sec:rep}
In this section, we introduce edge representation for spatial interpolation inspired by \cite{zhang2021representation}. 
The proposed edge representation  $e_{ji}$ for an edge $ (v_j, v_i)$, where $v_i$ is the target node and $v_j$ is the source node, can be expressed as
\begin{equation}\label{eq:rep_k}
        {e}_{ji}=\left (l_{ij} ,  \lambda_{ijk}\right),
\end{equation}
where $v_k$ is the neighbor of $v_j$ that forms the smallest $\lambda_{ijk}\in [-\pi, \pi)$, 
\begin{equation}\label{eq:rep2_k}
    \begin{split} 
     \lambda_{ijk}&=\mathrm{Parity}\cdot \bar{\lambda}_{ijk},\\
     \bar{\lambda}_{ijk}&=\mathrm{arccos} (\langle\frac{\mathbf{s}_{ij}}{l_{ij}},\frac{\mathbf{s}_{jk}}{l_{jk}}\rangle),\\
    l_{ij}&=\Vert \mathbf{s}_{ij} \Vert_{2} ,\\ 
     \mathbf{s}_{ij}&=\bar{s}_{j} - \bar{s}_{i},\\
     \mathrm{Parity}&=\langle\mathbf{n}_{ijk},\mathbf{n}_{xy}\rangle,\\
     \mathbf{n}_{ijk}&=\frac{\mathbf{s}_{ij} \times \mathbf{s}_{jk}}{\|\mathbf{s}_{ij}\times \mathbf{s}_{jk}\|_{2}},\\
     \mathbf{n}_{xy}&=\mathbf{u}_{x}\times\mathbf{u}_{y},
     \end{split}
\end{equation}
where $\bar{s}_{j}$ and $\bar{s}_{j}$ denote the coordinate values of two locations, $\mathbf{u}_{x}$ and $\mathbf{u}_{y}$ are unit vectors along the horizontal and vertical axis of the coordinate system on the interested plane, $\mathbf{n}_{xy}$ is the normal of the interested plane, $\times$ denotes the cross product operation.
\section{Experiment}
In this section, we first introduce the experimental settings, then we compared the effectiveness of the proposed model with comparison methods on ten real-world datasets.  All the experiments are conducted on a 64-bit machine with an NVIDIA A5000 GPU.
% Secondly, we study the sensitivity of hyperparameters and analyzed the efficiency of the proposed model. The experiment results indicate that our model achieved better or comparable performance on all data sets. 
\subsection{Experiment setting}
\paragraph{Dataset} 
We evaluate our method on ten real-world datasets, including seven civil unrest event prediction datasets and one influenza outbreak event prediction dataset extracted from Twitter data for the classification task, and two environmental datasets collected by in-situ monitoring sensors and satellites for the regression task. 
\begin{itemize}
\item \textbf{Civil unrest twitter datasets} Seven civil unrest event datasets from Brazil, Chile, Colombia, Ecuador, EI Salvador, Uruguay, and Venezuela are utilized to evaluate the performance of the proposed model. Details of these datasets could be found in \cite{zhao2015multi,muthiah2016embers}. 
\item \textbf{Influenza outbreak twitter dataset} Flu activities are collected from 48 states in the U.S. in this dataset. Details of these datasets could be found in \cite{zhao2017feature,zhao2015simnest,zhao2021event} We call this dataset Flu in the following sections. 
\item \textbf{PM2.5 concentration dataset} PM2.5 data in the Los Angeles region derived from the fusion of data collected by PurpleAir sensors and the Moderate Resolution Imaging Spectroradiometer (MODIS) TERRA and AQUA satellites \cite{levy2013collection}, as well as the meteorological dataset from MERRA-2 reanalysis data \cite{gelaro2017modern}. The dataset contains latitude, longitude,  and meteorological values such as humidity, surface pressure, wind speed, and the corresponding ambient PM2.5 value in the location.
\item \textbf{Ambient temperature dataset} In-situ air temperature was downloaded from Weather Underground, a network of weather stations. Satellite-based land surface temperature (LST) products derived from MODIS satellite observations and meteorological variables were collocated together to estimate ambient temperature.
\end{itemize}
\paragraph{Comparison method}
To the best of our knowledge, there has been little work handling unseen spatial domains. The following methods were included for comparing performances on the collected datasets.
\begin{itemize}
    \item \textbf{ERM}: A space-oblivious model which is trained on all training domains using ERM.
    \item \textbf{IncFinetune}: In this model, we incrementally train a global model on all training domains by finetuning the model on the training domains one at a time.
    \item \textbf{GTWNN}~\cite{feng2021geographically}: A geographically weighted neural network consisting of two artificial neural networks with the first network estimating the spatial weight of each independent variable from coordinate values.
\end{itemize}
\subsection{Experimental performance}
We adopt Area under the ROC Curve (AUC) score and mean absolute error (MAE) as the metrics for classification and regression tasks, respectively.
\begin{table*}
  \caption{Comparison of our proposed method against existing methods on all ten datasets in terms of MAE for the first two datasets and AUC score for others. The standard deviation over three runs follows the $\pm$ mark. We observe that our proposed method outperforms almost all the baselines}
% \begin{adjustbox}{width={\textwidth},keepaspectratio} %inside caption
    \centering
  \begin{tabular}{l|l|l|l|l|l}
%    {c|c|c|c|c|c|c|c}
    \toprule
    Dataset & ERM &IncFinetune &GTWNN  & SIGNN-G  & SIGNN\\
    \hline%\hline do not have line-space
    PM2.5 & 12.44 $\pm$ 4.64 &13.73 $\pm$ 4.07 &10.00 $\pm$ 0.58 &9.66 $\pm$ 0.48  & \textbf{9.40 $\pm$ 0.46}\\
    
    Temperature &8.74 $\pm$ 1.23 &11.13 $\pm$ 4.93 &12.29 $\pm$ 7.81 &7.41 $\pm$ 0.30  & \textbf{7.33 $\pm$ 0.28}\\
    \hline

    Flu & \textbf{0.84 $\pm$ 0.03} &0.80 $\pm$ 0.03 &0.75 $\pm$ 0.02 &0.74 $\pm$ 0.05  &\textbf{0.84} $\pm$ \textbf{0.06}\\

    Brazil  & 0.53 $\pm$ 0.03 &0.52 $\pm$ 0.03 &0.59 $\pm$ 0.04 &0.61 $\pm$ 0.08  &\textbf{0.65 $\pm$ 0.07}\\
    
    Chile  & 0.46 $\pm$ 0.04 &0.44 $\pm$ 0.12 &0.49 $\pm$ 0.07 &\textbf{0.57 $\pm$ 0.08}  &0.55 $\pm$ 0.05\\

    Columbia  & 0.52 $\pm$ 0.08 &0.44 $\pm$ 0.06 &0.55 $\pm$ 0.07 &\textbf{0.56 $\pm$ 0.04}  &\textbf{0.56 $\pm$ 0.11}\\
    
    Ecuador  & 0.47 $\pm$ 0.08 &0.38 $\pm$ 0.13 &0.47 $\pm$ 0.03 &\textbf{0.52 $\pm$ 0.08}  &\textbf{0.52 $\pm$ 0.18}\\
    
    El salvador  & 0.50 $\pm$ 0.07 &0.51 $\pm$ 0.08 &0.46 $\pm$ 0.07 &0.52 $\pm$ 0.07  &\textbf{0.53 $\pm$ 0.20}\\

    Uruguay  & 0.48 $\pm$ 0.08 &0.50 $\pm$ 0.10 &0.39 $\pm$ 0.12 &0.40 $\pm$ 0.17  &\textbf{0.54 $\pm$ 0.01}\\
    
    Venezuela  & 0.51 $\pm$ 0.03 &{0.55 $\pm$ 0.04} &0.56 $\pm$ 0.05 &\textbf{0.60 $\pm$ 0.03}  &0.54 $\pm$ 0.03\\
    
    \bottomrule
    \end{tabular}
    \label{tab:perf}
    % \end{adjustbox}
\end{table*}
\subsubsection{Effectiveness results}
Table \ref{tab:perf} summarizes the performance comparison among the proposed methods and competing models for civil unrest event forecasting, influenza outbreak prediction, ambient PM2.5 concentration, and temperature estimation tasks. The results show the proposed method achieves the best performance on most datasets and has comparable performance on other datasets. It indicates the method that adapts to different locations can better model the heterogeneous relationships among independent variables and dependent along the changes of locations. For example, for the seven civil unrest event dataset, the proposed model has the highest AUC scores in most countries except Venezuela. Specifically, the AUC scores of our model in Chile and Brazil are much higher than that of baseline models. 

\subsubsection{Ablation study}
We further conduct an ablation study on all ten datasets to evaluate the effectiveness of different components in our proposed model. Firstly, we remove the interpolation function in SIGNN and train a single global spatial embedding for all the locations and use this spatial embedding to generate the weights of the downstream-task model. We name this version of our method as SIGNN-G. The results of this version are included in Table \ref{tab:perf}.

As we can see, the interpolation function provided by SIGNN contributes significantly to the overall model performance. The difference in performance between SIGNN-G is an indicator of the extent of heterogeneity of the spatial data. This further implies that spatial heterogeneity exists in almost all the datasets except Columbia and Ecuador, on which the average performances of SIGNN and SIGNN-G are the same.
% \subsubsection{Efficiency analysis}
% To validate the efficiency of the proposed method, we record the average training time per epoch among all the models for 20 epochs.
% \subsubsection{Scalability}
% To test the scalability of the proposed method, we record the training time  versus the number of domains and the samples in each domain.
% \subsubsection{sensitivity analysis}
% We conduct sensitivity analysis on the number of neighbors $K$ chosen to build the KNN graph $G$.
% \subsection{Case Study}
% Visualization of the true PM2.5 and predicted
\section{Conclusion}
Spatial autocorrelation and spatial heterogeneity widely exist in spatial data, which makes the traditional machine learning model perform badly. Spatial domain generalization is a spatial extension of domain generalization, which can generalize to unseen spatial domains in continuous 2D space. Specifically, it learns a model under varying data distributions that generalizes to unseen domains.  Although tremendous success has been achieved in domain generalization, there exist very few works on spatial domain generalization. This paper proposes a generic framework for spatial domain generalization. Specifically, We develop a spatial interpolation graph neural network that handles spatial data as a graph and learns the spatial embedding on each node and their relationships. The spatial interpolation graph neural network infers the spatial embedding of an unseen location during the test phase. Then the spatial embedding of the target location is used to decode the
parameters of the downstream-task model directly on the target
location. Extensive experiments on ten real-world datasets demonstrate the proposed method's strength. SIGNN achieves the best performances on most of the datasets and comparable performance on the others. The difference in the performances on SIGNN-G and SIGNN validated our assumption that spatial heterogeneity exists in most spatial datasets.

\section*{Acknowledgment}
This work was supported by the National Science Foundation(NSF) Grant No. 1755850, No. 1841520, No. 2007716, No. 2007976, No. 1942594, No. 1907805, a Jeffress Memorial Trust Award, Amazon Research Award, NVIDIA GPU Grant, and Design Knowledge Company (subcontract number: 10827.002.120.04).
%\section*{References}
\bibliographystyle{IEEEtran}
\bibliography{IEEEabrv,ref}

\end{document}